\documentclass{article} % For LaTeX2e
\usepackage[preprint]{colm2026_conference}

\usepackage{microtype}
\usepackage{hyperref}
\usepackage{url}
\usepackage{booktabs}
\usepackage{algorithm}
\usepackage{algorithmic}
\usepackage{amsmath}
\usepackage{graphicx}
\usepackage[table]{xcolor}
\usepackage{subcaption}
\usepackage{wrapfig}

\usepackage[most]{tcolorbox}
\usepackage{fvextra}

\usepackage{lineno}

\definecolor{darkblue}{rgb}{0, 0, 0.5}
\hypersetup{colorlinks=true, citecolor=darkblue, linkcolor=darkblue, urlcolor=darkblue}

\title{Student-in-the-Loop Chain-of-Thought Distillation via \\ Generation-Time Selection}

\author{Chaoqun He, Yingfa Chen, Chaojun Xiao\thanks{Corresponding Authors.}, Xu Han, Lijie Wen\footnotemark[1]  \\
Tsinghua University \\
\texttt{hechaoqun1998@gmail.com}
}

\newcommand{\name}[0]{Gen-SSD}

\begin{document}

\ifcolmsubmission
\linenumbers
\fi

\maketitle

\begin{abstract}
% The abstract paragraph should be indented 1/2~inch (3~picas) on both left and
% right-hand margins. Use 10~point type, with a vertical spacing of 11~points.
% The word \textit{Abstract} must be centered and in point size 12. Two
% line spaces precede the abstract. The abstract must be limited to one
% paragraph.
% Large reasoning models achieve strong performance on complex reasoning tasks through long chain-of-thought (CoT) trajectories. Their massive scale makes them costly to deploy, motivating efforts to distill their reasoning ability into smaller models. However, direct long-CoT distillation often underperforms due to the capacity gap and the distribution mismatch between teacher and student models. 
% To address this issue, we propose \textbf{SSD} (\textbf{S}elf-\textbf{S}election \textbf{D}istillation), where the student actively selects suitable data during the teacher's sampling process.
% For each instruction, the student leverages perplexity scores to identify suitable continuations, enabling early pruning of unhelpful reasoning trajectories.
% Extensive experiments on mathematical reasoning benchmarks demonstrate that SSD outperforms standard knowledge distillation by an average of more than 6 points and other baseline methods. 
% Ablation studies confirm the robustness and generality of SSD under different teacher scales and closed-source settings. Our results demonstrate that effective data selection is the key to distilling reasoning ability for small models.
Large reasoning models achieve strong performance on complex tasks through long chain-of-thought (CoT) trajectories, but directly transferring such reasoning processes to smaller models remains challenging. A key difficulty is that not all teacher-generated reasoning trajectories are suitable for student learning.
Existing approaches typically rely on post-hoc filtering, selecting trajectories after full generation based on heuristic criteria. However, such methods cannot control the generation process itself and may still produce reasoning paths that lie outside the student’s learning capacity.
To address this limitation, we propose 
\textbf{Gen-SSD} (\textbf{Gen}eration-time \textbf{S}elf-\textbf{S}election \textbf{D}istillation)
, a student-in-the-loop framework that performs generation-time selection. Instead of passively consuming complete trajectories, the student evaluates candidate continuations during the teacher’s sampling process, guiding the expansion of only learnable reasoning paths and enabling early pruning of unhelpful branches.
Experiments on mathematical reasoning benchmarks demonstrate that Gen-SSD consistently outperforms standard knowledge distillation and recent baselines, with improvements of around 5.9 points over Standard KD and up to 4.7 points over other baselines.
Further analysis shows that Gen-SSD produces more stable and learnable reasoning trajectories, highlighting the importance of incorporating supervision during generation for effective distillation.
\end{abstract}

\section{Introduction}
Recent advances in large reasoning models (LRMs), such as DeepSeek-R1~\citep{deepseekai2025deepseekr1incentivizingreasoningcapability} and OpenAI-o1/o3~\citep{openai2024openaio1card}, have led to remarkable progress on complex reasoning tasks, such as mathematical problem solving~\citep{lightman2023let,besta2024graph,muennighoff2025s1} and code generation~\citep{jiang2024survey}. 
The key to this success is the use of long chain-of-thought~(CoT)~\citep{wei2022chain,li202512surveyreasoning}, where intermediate reasoning steps are explicitly generated before arriving at the final answer.

% \begin{figure}[!t]
%     \centering
%     \includegraphics[width=0.5\linewidth]{pics/intro.pdf}
%     \caption{Comparison between standard knowledge distillation and our proposed SSD. Standard KD trains the student on all teacher outputs indiscriminately, while SSD enables the student to actively select suitable data during the teacher's sampling process, leading to improved performance.}
%     \label{fig:intro}
% \end{figure}

\begin{wrapfigure}{r}{0.4\linewidth}
    \centering
    \includegraphics[width=\linewidth]{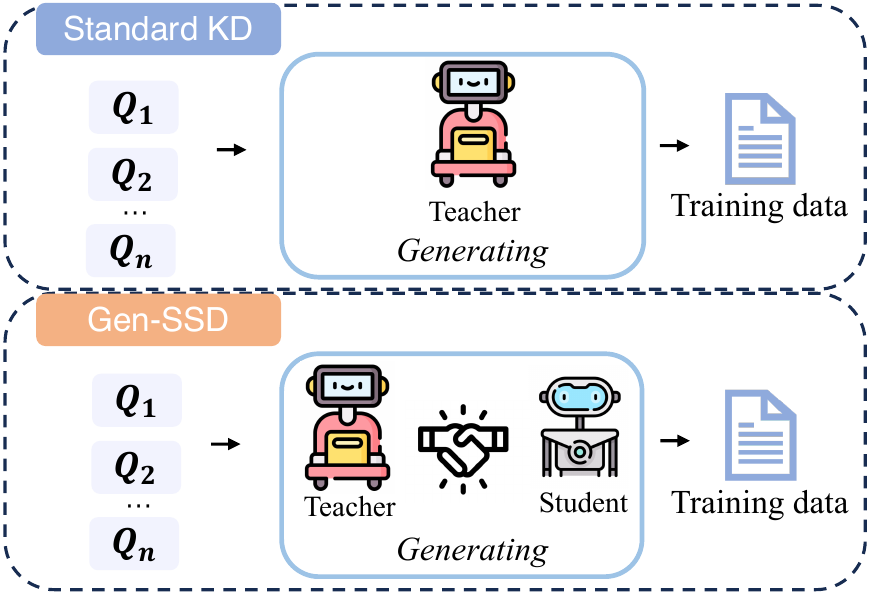}
    \caption{Comparison between standard KD and our proposed \name{}. }
    % Standard KD trains the student on all teacher outputs indiscriminately, while SSD enables the student to actively select suitable data during the teacher's sampling process, leading to improved performance.}
    \label{fig:intro}
\end{wrapfigure}

However, the superior performance of LRMs comes at a steep cost: their massive parameter sizes demand expensive computational resources.
% , making them inaccessible to most academic researchers and small-scale enterprises. 
A natural alternative is to deploy smaller language models that approximate the reasoning capabilities of LRMs while being computationally affordable. Knowledge distillation~\citep{hinton2015distilling,kim2016sequence,agarwal2024policy} has emerged as a promising strategy, where a strong teacher model transfers its reasoning ability to a weaker student model by supervising it with long CoT data.

% Despite its appeal, recent studies have found that standard knowledge distillation~(Standard KD)~\citep{ho2023large}---as shown in Figure~\ref{fig:intro}, where long CoT trajectories are directly distilled from large teacher models to smaller student models---often fails to deliver the expected improvements~\citep{li-etal-2025-small-models,yin2025towards}.
% The primary challenge lies in the capacity gap between teacher and student: small models struggle to consistently benefit from complex, lengthy reasoning trajectories~\citep{zhang-etal-2023-lifting}. For instance, \citet{li-etal-2025-small-models} observe that students trained on short and simple CoT achieved larger improvements than those distilled purely from long and complex CoT.
% \citet{ding2025micota} introduce an intermediate teacher assistant and employ intermediate-length CoT data to mitigate the capacity gap. 
% Thus, we observe that not all data generated by the teacher should be directly used for training; effective distillation requires selecting a data subset that the student model can effectively learn from.
% Yet, existing approaches rely on manually constructed heuristics for data generation, limiting their automation and scalability. This raises a question: \emph{can we leverage the student itself to automatically identify and select the data most suitable for its own learning?}

Despite its appeal, recent studies have found that standard knowledge distillation~(Standard KD)~\citep{ho2023large}—as illustrated in Figure~\ref{fig:intro}, where long CoT trajectories are directly transferred from large teacher models to smaller student models—often fails to deliver the expected improvements~\citep{li-etal-2025-small-models,yin2025towards}. A key observation is that not all teacher-generated reasoning trajectories are equally useful for student learning. In particular, small models may struggle to benefit from long and complex reasoning processes, and prior work has shown that shorter or simpler CoT can sometimes lead to better performance~\citep{li-etal-2025-small-models}. 
% To address this, some methods introduce intermediate supervision, such as teacher assistants or simplified reasoning trajectories~\citep{ding2025micota}.
To address this, some prior work has explored post-hoc filtering of teacher-generated data~\citep{chen2023mcc,yan2025efficientcotdistillationselfguided}, as well as the use of intermediate supervision, such as teacher assistants or simplified reasoning trajectories~\citep{ding2025micota}.
These results suggest that the effectiveness of distillation depends on selecting training data that the student can actually learn from. 
% However, existing approaches typically rely on manually designed heuristics to construct such data, which limits their flexibility and scalability. 
However, existing approaches typically perform selection after the trajectories have been fully generated, relying on heuristic rules to filter the data. 
As a result, the learnability of trajectories is fixed once they are generated and cannot be adjusted afterward.
% As a result, the learnability of the trajectories is already fixed, and cannot be adjusted during the generation process.
% This naturally raises the question: \emph{can the student model itself be used to identify and select reasoning trajectories that are more suitable for its own learning?}
This raises a natural question: \emph{can the student be involved during generation to guide the selection of reasoning trajectories that are more suitable for its own learning?}

In this work, we propose a generation-time self-selection framework for CoT distillation, \name{}. 
% Specifically, we introduce \textbf{\name{}} (\textbf{Gen}eration-time \textbf{S}elf-\textbf{S}election \textbf{D}istillation), 
We allow the student to actively participate in the teacher's data generation process. 
As illustrated in Figure~\ref{fig:method}, instead of passively receiving all teacher outputs, the student evaluates partial sequences using its own perplexity (PPL) and selects prefixes that are compatible with the student's capacity. By intervening early at the generation time, \name{} prunes unhelpful candidates, reduces computational overhead, and tailors the training data to the student's capabilities.

% Extensive experiments demonstrate the effectiveness of our approach. Using QwQ-32B~\citep{qwq32b} as the teacher model and Qwen2.5-Math-1.5B~\citep{yang2024qwen2} as the student model, \name{} achieves superior performance over other baselines across the benchmarks.
Extensive experiments demonstrate the effectiveness of our approach. Using QwQ-32B~\citep{qwq32b} as the teacher model and Qwen2.5-Math-1.5B~\citep{yang2024qwen2} as the student model, \name{} consistently outperforms baselines across a range of math reasoning benchmarks, with improvements of around 5.9 points over Standard KD and up to 4.7 points over other baselines. The improvements are particularly evident on tasks that require multi-step reasoning, suggesting that our method provides more suitable supervision for learning structured reasoning processes.
% Ablation studies further demonstrate the generality and robustness of \name{}, underscoring the importance of data selection in reasoning distillation. We will release our code and data to facilitate future research.
Ablation studies further demonstrate the generality and robustness of \name{}. 
% In particular, we compare different PPL-based selection strategies (low, high, and random), and observe that selecting low-PPL chunks consistently leads to better performance, highlighting the importance of selecting learnable supervision. These results underscore the importance of data selection in reasoning distillation. 
In particular, we compare different PPL-based selection strategies (low, high, and random) and observe clear differences in performance. Among them, low-PPL selection generally performs better, suggesting that the way training data is selected plays a critical role, and that not all trajectories are equally suitable for the student to learn from.
We will release our code and data to facilitate future research.

\section{Related Work}

\subsection{Knowledge Distillation}
Traditional knowledge distillation approaches rely on logits distillation, where the student is trained to match the teacher's output probability distributions~\citep{hinton2015distilling,beyer2022knowledge,sanh2019distilbert}. 
% With the rise of large language models, two categories of KD are commonly applied: black-box distillation, where only model outputs are accessible~\citep{kim2016sequence,taori2023stanford,wang-etal-2023-self-instruct}, and white-box distillation, which leveraging soft labels from teacher models~\citep{gu2023minillm,yang2024survey}.
With the rise of large language models, two categories of KD are commonly applied: black-box distillation, where only model outputs are accessible~\citep{kim2016sequence,taori2023stanford,wang-etal-2023-self-instruct}, and white-box distillation, which leverages soft labels from teacher models~\citep{gu2023minillm,yang2024survey}. In practice, black-box methods are more widely applicable due to limited access to internal model states, while white-box methods can provide richer supervision when such access is available.

With the introduction of CoT~\citep{wei2022chain}, researchers have sought to distill not only the final predictions but also the reasoning processes of teacher models into smaller students~\citep{deepseekai2025deepseekr1incentivizingreasoningcapability,faceopen}. The hope is that students can learn reasoning skills by imitating teacher-generated CoT trajectories. However, several studies~\citep{li-etal-2025-small-models,ding2025micota} have shown that direct distillation of trajectories is often suboptimal, as smaller models struggle to fully absorb complex trajectories from much stronger teachers.

Recently, on-policy distillation has attracted increasing attention, as it leverages dense distillation rewards and reinforcement learning–based self-exploration to improve student model performance~\citep{gu2023minillm,agarwal2024policy,lu2025onpolicydistillation,yang2025qwen3,mimo2025flash}. 
Such methods typically rely on a strong initialization for student models to facilitate effective subsequent exploration. Our approach can be naturally applied at the cold-start stage, helping to establish a stronger initialization for on-policy distillation.

\subsection{Data Engineering for CoT Distillation}

\citet{ho2023large} propose distilling CoT trajectories from teacher models into smaller student models, alleviating the limitation that many state-of-the-art models are closed-source. Building on this direction, \citet{zhang2024elad} enhance knowledge transfer through active learning and explanation-guided sample selection. Subsequent studies~\citep{muennighoff2025s1,ye2025limo} focus on improving sample quality by selecting a smaller but more informative subset of training data, while others aim to refine the reasoning trajectories.

% Among these methods, 
~\citet{chen2023mcc} improve reasoning consistency by generating multiple trajectories per question and minimizing the bidirectional KL divergence between their corresponding answer distributions. Similarly, \citet{li-etal-2025-small-models} propose mixing short CoT generated by instruction-tuned models with long CoT from LRMs to enhance distillation effectiveness. \citet{zhou2024teaching,ding2025micota} introduce a teacher assistant model to facilitate more suitable data selection. However, these methods primarily rely on heuristic criteria without explicitly accounting for the student’s learning capacity. MoRSD~\citep{yan2025efficientcotdistillationselfguided} incorporates a difficulty-aware metric for trajectory selection. However, its diversity-based filtering may inadvertently discard student-aligned reasoning trajectories, limiting its effectiveness.

% 强调这些相关方法的post-hoc selection
Despite their differences, these methods generally follow a similar paradigm: they operate on completed trajectories and perform selection or filtering only after the full reasoning process has been generated. As a result, the structure and difficulty of these trajectories are already determined, and the generation process itself is not directly controlled. While such post-hoc selection can help reduce some misaligned supervision, it is still limited in its ability to avoid generating unlearnable or suboptimal reasoning paths in the first place.

\section{Method}

\begin{figure*}[htbp]
    \centering
    \includegraphics[width=1\linewidth]{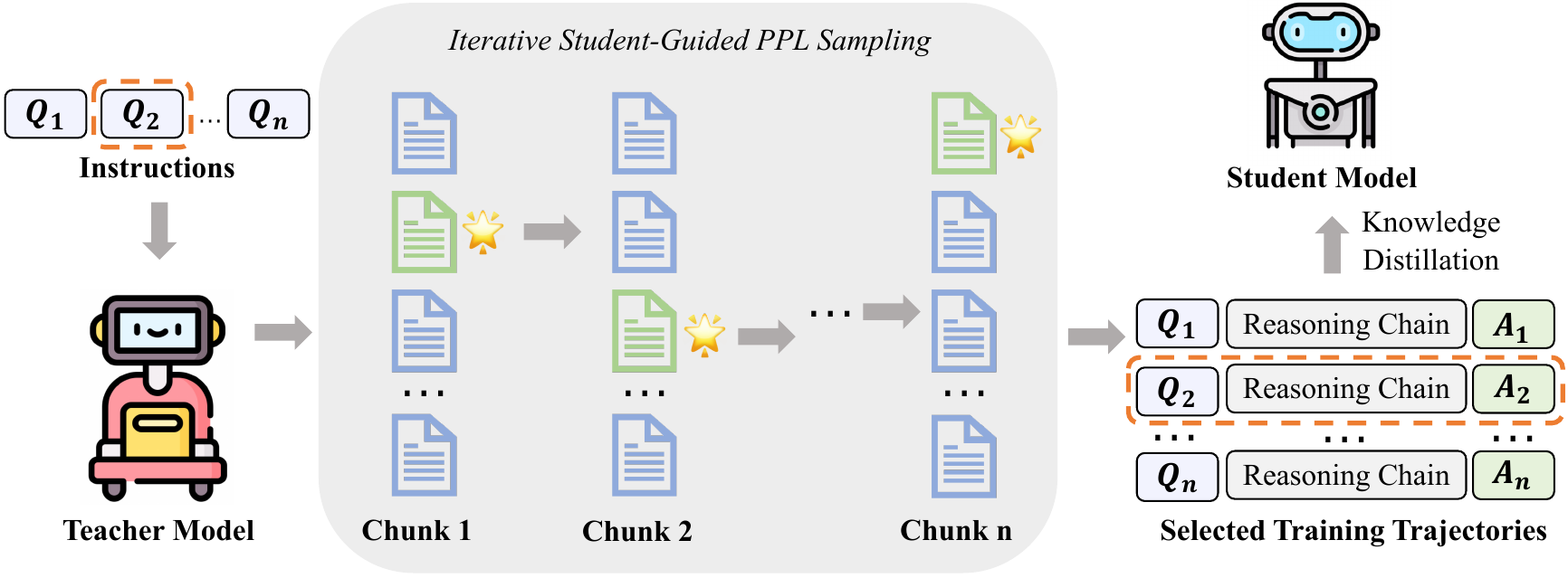}
    \caption{Overview of \name{}. The student actively participates in the teacher's multi-sample generation process. At each chunk, the student evaluates candidate continuations with PPL and selects the fragments best aligned with its capability, thereby influencing the teacher's sampling trajectory. For unsuitable candidates, generation is terminated early, which reduces inference cost and improves sampling efficiency.}% In the second stage (Training), the selected data are further refined via rejection sampling and used to fine-tune the student model.}
    \label{fig:method}
\end{figure*}

In this section, we describe the core components of \name{} in detail. Figure~\ref{fig:method} illustrates our proposed \name{}. Given a set of problems, the teacher generates multiple reasoning candidates chunk by chunk. At each step, the student evaluates the candidates using PPL and selects the continuation that matches its capacity. This self-selection mechanism allows the student to guide the teacher's sampling trajectory, resulting in training trajectories that are both aligned with the student's capacity and cost-efficient. We next describe the core components of \name{} in detail. We first present the problem formulation, followed by the overall data selection process, with the full procedure summarized in Algorithm~\ref{alg:ssd}. We then explain the rationale for using PPL as the selection criterion. 

\subsection{Problem Setup}

The goal of knowledge distillation is to transfer the ability of a strong teacher model $T$ into a smaller student model $S$. Formally, given a set of problems
$\mathcal{X} = \{x_1, x_2, \dots, x_n\}$,
the teacher generates reasoning trajectories and final answers:
$y_i \sim T(\cdot \mid x_i)$,
which are used as supervision for supervised fine-tuning of the student:
\begin{equation}
\min_{\theta_S} \; \mathbb{E}_{(x,y)\in\mathcal{D}_T} \big[-\log P_S(y \mid x; \theta_S)\big],
\end{equation}
where $\mathcal{D}_T$ denotes the teacher-generated dataset. This standard pipeline passively accepts the teacher's outputs, regardless of whether they are aligned with the student's capacity.

In contrast, our proposed \name{} modifies the sampling process itself by incorporating the student into the loop. Specifically, instead of blindly adopting all teacher generations, \name{} allows the student to evaluate candidates during the teacher's multi-sample generation. At each chunk, the student selects the continuation with lower PPL, thereby guiding the teacher's sampling trajectory and retaining only trajectories that the student can effectively learn from. The selected trajectories are then used in the SFT stage as in standard distillation.

\subsection{Data Selection}

% A central challenge in reasoning distillation is the capacity gap between the teacher and student models~\citep{ding2025micota}.  
% Standard KD ignores this gap, forcing the student to imitate teacher's outputs indiscriminately. As a result, the student may fail to benefit from teacher trajectories that are too complex or misaligned with its distribution.
A central challenge in reasoning distillation is that not all teacher-generated trajectories are equally useful for student learning~\citep{ding2025micota}.  
Standard KD treats all teacher outputs in the same way, without distinguishing whether they are actually helpful for the student. As a result, the student may struggle to benefit from trajectories that are overly complex or not well-matched to their current capabilities.

To address this issue, we introduce a student-in-the-loop selection mechanism. Instead of passively accepting the teacher's outputs, the student actively evaluates teacher-generated candidates during generation and selects those most compatible with its own capabilities. This ensures that the distilled data are not only correct but also learnable for the student.

\textbf{Cold Start.}
Before applying self-selection, we introduce a lightweight cold-start phase to bootstrap the student.
This phase aligns the student with the teacher’s reasoning format and enables the student to produce more meaningful signals for selecting teacher-generated reasoning candidates.
QwQ-32B\footnote{\url{https://huggingface.co/Qwen/QwQ-32B}} is trained to generate outputs containing \texttt{<think>}, \texttt{</think>} tokens that explicitly delimit reasoning chains, 
we first teach the student model to follow the format.
Thus, we use a few reasoning chains of teacher models to initialize the student model. This step ensures format alignment between the teacher and student. 
% Moreover, \citet{li-etal-2025-small-models} has shown that base models suffer from a larger capability gap compared to instruct models, which motivates us to apply a cold-start phase to partially mitigate this gap.
Moreover, \citet{li-etal-2025-small-models} show that base models are less able to directly benefit from such supervision compared to instruction-tuned models, which motivates us to include a cold-start phase to make the training signals more accessible to the student.

Specifically, given candidate dataset $\mathcal{D}_{\text{train}}$, we sample from $T$ with rejection sampling to collect a small set of verified reasoning trajectories $\mathcal{D}_{\text{init}}$:
\begin{equation}
\begin{aligned}
% \mathcal{D}_{\text{init}} = \{ (x, y) \mid y \sim T(\cdot \mid x), \; & \\ \text{Answer}(y) = \text{Answer}(x) \}.
\mathcal{D}_{\text{init}} = \{(x, \hat{y}) \mid \text{Answer}(\hat{y}) = y,\; \hat{y} \sim T(x),\; (x, y) \sim \mathcal{D}_{\text{train}}\}
\end{aligned}
\end{equation}

The student $S$ is fine-tuned on $\mathcal{D}_{\text{init}}$ to better align its distribution with that of the teacher before Self-Selection.

\textbf{Self-Selection.}
To enable capacity-aware distillation, the student model is allowed to actively select reasoning candidates that match its own learning ability.
During multi-sample generation by the teacher, after producing each chunk of tokens, the student evaluates the candidate continuations. Using PPL as the selection metric, the student chooses the candidate with the lower PPL, which is used to continue subsequent sampling until termination. 
This early intervention effectively prunes unsuitable sequences, avoiding wasted computation on data that the student can't learn from. 

Specifically, the teacher generates text in chunks of length $m$. At chunk step $c$, the teacher produces $K_c$ candidates:
\begin{equation}
\begin{aligned}
\mathcal{Y}c = \left\{ y_c^{(1)}, y_c^{(2)}, \dots, y_c^{(K_c)} \right\}, \quad y_c^{(k)} \sim T(\cdot \mid x, y_{<c}),
\end{aligned}
\end{equation}
where $y_{<c}$ denotes the concatenation of previously generated chunks.

The student then evaluates each candidate by computing PPL. Then, the candidate with the lowest PPL is selected as the continuation:
\begin{equation}
y_c^\ast = \arg\min_{y_c^{(k)} \in \mathcal{Y}_c} \text{PPL} \left(y_c^{(k)} \right).
\end{equation}
The selected chunk is concatenated with the preceding context and then fed back to the teacher for generating the next chunk.

To further improve efficiency and reduce inference cost, we progressively decrease the number of samples per chunk:
% \begin{equation}
% \label{eq:sampling}
% K_1 = 16, \quad K_2 = 8, \quad K_c = 4 \;\;\; \text{for } c \geq 3.
% \end{equation}
\( K_1 = 16,\; K_2 = 8,\; K_c = 4 \text{ for } c \ge 3 \).
% This adaptive strategy lowers computation while preserving data quality.
As validated by our experiments in Appendix~\ref{app:sampling_decrease}, compared to consistently sampling 16 candidates, this adaptive strategy reduces sampling time while preserving data quality.

After data selection, only one trajectory $y^\ast = (y_1^\ast, \dots, y_C^\ast)$ is retained for each problem, specifically the one with the lowest PPL. 
% Then we apply rejection sampling by keeping only correct solutions:
In constructing the final training set, only reasoning trajectories that yield correct answers are preserved:
\begin{equation}
\begin{aligned}
\mathcal{D}_{\text{SSD}} = \{ (x, y^\ast) \mid \text{Answer}(y^\ast) = y,\; (x, y) \sim \mathcal{D}_{\text{train}}\}
\end{aligned}
\end{equation}

\subsection{Why we choose Low-PPL}
% Perplexity as a Student-Aware Selection Signal
\name{} adopts PPL as a student-aware signal for trajectory selection, focusing on identifying trajectories that align with the student’s learning capacity rather than measuring absolute reasoning quality. In reasoning distillation, selecting learnable trajectories is often more effective than exposing the student to overly complex trajectories. Consistent with this intuition, our comparison of low/high/random-PPL selected data under identical settings in Table~\ref{tab:low-high-random-ppl_exp} shows that training on low-PPL data consistently yields better performance, empirically validating the effectiveness of PPL-based selection in \name{}.

\section{Experiments}
% 主实验有：最好的span-level、self-student、baseline。
% ablation：不同的chunk-size，有无冷启动

In this section, we first describe the experimental details, including the training data, evaluation benchmarks, models, baselines and training configurations~(Section~\ref{sec:setup}). Then, we present the main results (Section~\ref{sec:main-results}).
Finally, we provide a detailed analysis of why \name{} outperforms other methods (Section~\ref{sec:analysis}).
% the effectiveness of our method when distilling from closed-source APIs (Section~\ref{sec:application-to-closed-source}). 
% Finally, we conduct further analysis on the ablation of critical design choices and the average PPL of different data generation methods (Section~\ref{sec:ablation-studies}).

\subsection{Setup}
\label{sec:setup}

\textbf{Training Data.}
OpenMathReasoning~\citep{moshkov2025aimo2} is a large-scale math reasoning dataset, which contains 540K unique mathematical problems sourced from AoPS forums\footnote{\url{https://artofproblemsolving.com/community}}. They use Qwen2.5-32B-Instruct~\citep{qwen2.5} to preprocess problems, and DeepSeek-R1 and QwQ-32B to generate solutions. We build our training corpus based on it. Due to the large size of the original dataset, we construct a subset for our experiments. Specifically, we select problems for which DeepSeek-R1 generated 16 distinct solutions after deduplication and merging. In total, we retain 25K instructions, which are used in all subsequent experiments.

In the cold-start stage, we perform rejection sampling with QwQ-32B and retain 3K correct samples. In the subsequent Self-Selection stage, we apply our method to the remaining data and select 8.5K correct samples for training. To prevent data leakage, the two datasets are kept strictly disjoint. We follow the official recommended hyperparameters\footnote{\url{https://huggingface.co/Qwen/QwQ-32B}}, setting temperature = 0.6, top-$p$ = 0.95, top-$k$ = 30, and the maximum generation tokens = 16K.

\begin{table*}
  \centering
  \small
  \begin{tabular}{l|ccccc|c}
    \toprule
    Method & AIME25 & AIME24 & AMC23 & OlympiadBench & GSM8K & Avg $\uparrow$ \\
    % \textbf{Method}\rule{0pt}{3ex} & \textbf{AIME25} & \textbf{AIME24} & \textbf{AMC23} & \makecell{\textbf{Olympiad} \\ \textbf{Bench}} & \textbf{MATH500} & \textbf{GSM8K} & \textbf{Average} $\uparrow$ \\
    \midrule
    \rowcolor{gray!20}
    Teacher & 49.79 & 67.50 & 91.09 & 66.91 & 94.20 & 73.90 \\
    \rowcolor{gray!20}
    Student & 0.83  & 3.33  & 19.06 & 15.73 & 38.74 & 15.54 \\
    \midrule
    % base & & & & & & & \\
    % Standard KD* & 4.37 & 3.75 & 33.28 & 23.29 & 69.52 & 26.84 \\
    % MCC-KD* & 4.58 & 5.42 & 34.22 & 27.74 & 65.00 & 74.68 & 35.27 \\
    % MoRSD* & 7.92 & 6.88 & 37.50 & 28.64 & 64.80 & 76.27 & 37.00 \\ 
    % \name{}* & 6.25 & 6.88 & 34.84 & 28.64 & 68.84 & 29.09 \\
    % \midrule
    Cold Start & 3.75 & 0.42 & 19.53 & 16.17 & 51.63 & 18.30 \\
    Standard KD & 7.08 & 7.08 & 35.00 & 24.78 & 71.11 & 29.01 \\
    Self-Distillation & 2.71 & 2.71 & 29.53 & 22.55 & 63.76 & 24.25 \\
    MCC-KD & 5.21 & 5.63 & 35.47 & 27.15 & 77.79 & 30.25 \\
    MoRSD & 7.50 & 7.50 & 38.75 & \textbf{31.75} & 79.83 & 33.07 \\
    \name{} (ours) & \textbf{10.00} & \textbf{10.21} & \textbf{41.88} & 30.71 & \textbf{81.88} & \textbf{34.94}\\
    % \name{} (ours) & \textbf{10.00} & \textbf{8.96} & \textbf{41.88} & \textbf{30.71} & \textbf{66.40} & \textbf{79.98} & \textbf{39.66} \\
    \bottomrule
  \end{tabular}
  \caption{Performance comparison across various benchmarks. We adopt QwQ-32B as the teacher model and Qwen2.5-Math-1.5B as the student model. Best results are bolded. 
  The chunk size for the \name{} is 4K. All models are evaluated in a zero-shot setting, except for the student model, which is evaluated with two-shot prompting.}
  \label{tab:main_exp}
\end{table*}

\textbf{Evaluation Benchmarks.}
We evaluate our approach on a diverse set of mathematical reasoning benchmarks: AIME25~\citep{mathai2025}, AIME24~\citep{mathai2024}, AMC2023, OlympiadBench~\citep{he-etal-2024-olympiadbench}, and GSM8K~\citep{cobbe2021training}, covering a range of difficulty levels and reasoning styles.
For AIME25, AIME24, and AMC2023, we report Avg@16 due to the limited number of problems.
For OlympiadBench\footnote{\url{https://huggingface.co/datasets/Hothan/OlympiadBench}} and GSM8K, we report zero-shot Pass@1 accuracy.

\textbf{Models.}
For the teacher model, we choose QwQ-32B~\citep{qwq32b}, a strong LRM capable of producing long CoT trajectories.
For the student model, we use Qwen2.5-Math-1.5B\footnote{\url{https://huggingface.co/Qwen/Qwen2.5-Math-1.5B}}~\citep{yang2024qwen2}. To enable more accurate PPL estimation during self-selection, we perform a cold-start fine-tuning step on the base model using a subset of long CoT data from the teacher, ensuring the student is sufficiently aligned with the teacher's distribution.

% 这里也可以提到OMR的数据，因为QwQ-32B和R1可以验证我们方法的泛用性。
\textbf{Baseline Methods.}
We compare against four baseline methods:
% \begin{itemize}
(1)~\textbf{Standard KD}~\citep{ho2023large}. Direct distillation from the teacher's long CoT sequences via rejection sampling. To ensure fairness, we evaluate both the base student and the cold-started student under this setting.
(2)~\textbf{Self-Distillation}. The student performs rejection sampling on their own generated data. In this case, we use the cold-started student as the initial model.
% \end{itemize}
(3)~\textbf{MCC-KD}~\citep{chen2023mcc}, which enhances reasoning consistency by producing multiple trajectories per question and aligning their answer distributions via bidirectional KL divergence
(4)~\textbf{MoRSD}~\citep{yan2025efficientcotdistillationselfguided}, which distills CoT using a self-guided trajectory difficulty metric to select high-quality trajectories for efficient student training. We apply MCC-KD and MoRSD to the cold-started model setting.

\textbf{Implementation Details.}
We implement all experiments using the HuggingFace Transformers library and train on a server equipped with 4 NVIDIA A800-SXM4-80GB GPUs. The hyperparameters and training configurations are summarized in Table~\ref{tab:hyper-setting}.

\subsection{Main Results}
\label{sec:main-results}

We compare \name{} against the above baselines across all benchmarks, and the results are summarized in Table~\ref{tab:main_exp}. 
Overall, \name{} achieves the best performance across all benchmarks, outperforming Standard KD, MCC-KD, and MoRSD. In particular, \name{} improves over Standard KD and MCC-KD by 5.9 and 4.7 points, respectively, 
demonstrating the effectiveness of our method as well as the importance of allowing the student model to select data aligned with its own capacity.
While MoRSD uses a trajectory difficulty metric, its diversity-based selection may randomly exclude student-aligned trajectories, leading to limited improvements in mathematical reasoning scenarios. Furthermore, we extend our evaluation to code generation tasks and additional reasoning benchmarks. The results in Table~\ref{tab:generalizability_exp} show that \name{} maintains superior performance over the baselines, highlighting its robustness and generalizability.

\subsection{Analysis of Learnability and Trajectory Structure}
\label{sec:analysis}
% To understand why \name{} leads to improved performance compared to post-hoc filtering methods such as MoRSD, we analyze the selected reasoning trajectories from the perspective of student learnability.
To understand why \name{} leads to improved performance compared to post-hoc filtering methods such as MoRSD, which select trajectories only after generation is complete, we analyze the selected reasoning trajectories from the perspective of student learnability.

% First, we measure the average perplexity (PPL) of the student model on the selected trajectories. We find that SSD produces significantly lower PPL (1.5665) compared to MoRSD (2.4616), indicating that SSD generates supervision that is better aligned with the student’s distribution. In addition, SSD yields substantially shorter reasoning trajectories (5049.6 tokens on average) compared to MoRSD (6139.1 tokens), suggesting that generation-time selection avoids unnecessary or uninformative reasoning steps.
% First, we measure the average PPL of the student model on the selected trajectories. 
First, we measure the average PPL of the student and token length on trajectories generated for 1,000 randomly sampled problems.
As shown in Table~\ref{tab:ssd_morsd}, \name{} results in lower PPL than MoRSD, suggesting that the selected trajectories are more suitable for students to learn from. 
In addition, \name{} produces shorter reasoning trajectories on average, reducing the length by around 1,000 tokens, indicating that intervening during generation helps avoid unnecessary or uninformative steps, while also reducing computational overhead.

\begin{wraptable}{r}{0.4\linewidth}
    \centering
    \small
      \begin{tabular}{l|cc}
        \toprule
        Method & PPL $\downarrow$ & Token length $\downarrow$\\
        \midrule
        MoRSD & 2.46 & 6139.10 \\
        \name{} & \textbf{1.57} & \textbf{5049.60} \\
        \bottomrule
      \end{tabular}
    \caption{Comparison between \name{} and MoRSD in terms of student PPL and token length.}
    \label{tab:ssd_morsd}
\end{wraptable}
% We further examine the per-chunk PPL dynamics on a representative example (chunk size = 256). As shown in Figure~\ref{fig:ppl-trend}, trajectories selected by MoRSD exhibit large fluctuations in PPL, with several sharp spikes indicating locally unlearnable segments. In contrast, SSD produces much smoother PPL profiles, suggesting that it avoids abrupt distribution shifts during trajectory construction.
We further examine the per-chunk PPL along a representative example (chunk size = 256). As shown in Figure~\ref{fig:ppl-trend}, the PPL of trajectories selected by MoRSD varies more noticeably, with several sharp increases. In contrast, the trajectories produced by \name{} show a smoother pattern, suggesting a more stable progression during generation.

% A qualitative analysis reveals that high-PPL segments in MoRSD often correspond to exploratory reasoning patterns, characterized by speculative language, repeated hypothesis revisions, and non-convergent reasoning paths. These segments are difficult for the student model to learn due to their inconsistency and distributional mismatch. In contrast, SSD-selected trajectories exhibit structured and deterministic reasoning, with clear step-by-step decomposition and stable token patterns.
We further examine a specific chunk (index = 26), where the PPL difference is particularly pronounced. The corresponding text is shown in Section~\ref{sec:chunk_analysis}. At this point, the trajectory selected by \name{} is already close to completion, while MoRSD continues to generate without clear convergence, leading to unnecessarily long and exploratory reasoning.
A qualitative analysis shows that high-PPL segments in MoRSD often involve speculative language, repeated revisions, and non-convergent reasoning paths, which are harder for the student to learn. In contrast, \name{} produces more structured and direct reasoning, with clearer steps and more stable token patterns.

% These results suggest that SSD does not merely filter trajectories after generation, but actively shapes the reasoning process to remain within the student’s learnable region. This leads to more stable and effective supervision signals, explaining the empirical performance gains over post-hoc selection methods.
These results suggest that the advantage of \name{} over post-hoc selection methods such as MoRSD lies in when the student signal is used. Instead of filtering completed trajectories after generation, \name{} uses the student signal during generation, which helps avoid unsuitable reasoning paths early on. This leads to trajectories that are more stable and easier for the student to learn from, and likely contributes to the better distillation performance.

\begin{table*}
  \centering
  \small
  \begin{tabular}{l|ccccc|c}
    \toprule
    Method & AIME25 & AIME24 & AMC23 & OlympiadBench & GSM8K & Avg $\uparrow$ \\
    \midrule
    Cold Start & 3.75 & 0.42 & 19.53 & 16.17 & 51.63 & 18.30 \\
    \midrule
    w/o cold-start & 6.25 & 6.88 & 34.84 & 28.64 & 68.84 & 29.09 \\
    w/ cold-start & \textbf{10.00} & \textbf{10.21} & \textbf{41.88} & 30.71 & \textbf{81.88} & \textbf{34.94}\\
    \bottomrule
  \end{tabular}
  \caption{Comparison under settings with and without cold-start initialization.}
  \label{tab:cold-start}
\end{table*}

\section{Ablation Studies}
\label{sec:ablation-studies}

To better understand the effectiveness of our proposed \name{} framework, we conduct a series of ablation studies. Specifically, we examine the role of the cold-start stage, the impact of chunk size, the influence of different teacher model sizes, and the validity of PPL-based selection. These analyses provide deeper insights into the stability, generality, and efficiency of \name{} across varying configurations and settings.

\subsection{Effect of Cold Start}

We validate the importance of the cold-start stage. As shown in Table~\ref{tab:cold-start}, results show that 
% even without cold start, our proposed \name{} still outperforms standard KD, demonstrating the effectiveness of chunk-level data selection. Moreover, 
\name{} with cold start achieves an additional gain of over 5.8 points compared to \name{} without cold start. 
% This confirms that cold start indeed helps reduce the capability gap between teacher and student models, thereby providing a stronger foundation for subsequent distillation.
This suggests that the cold-start stage helps the student better adapt to the teacher’s reasoning format, providing a more stable starting point for distillation.

\subsection{Effect of Chunk Size}
\label{sec:chunk-size}

\begin{wrapfigure}{r}{0.55\linewidth}
    \centering
    \includegraphics[width=0.95\linewidth]{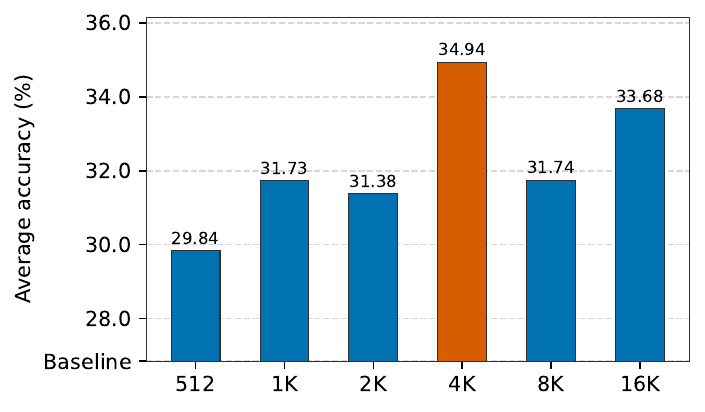}
    \caption{
    % Performance of \name{} with different chunk sizes across benchmarks, along with the overall average. The detailed results are provided in the Table~\ref{tab:benchmark_exp} in the appendix.
    Average performance of \name{} across benchmarks under different chunk sizes. Detailed results are provided in Table~\ref{tab:benchmark_exp} in the appendix.
    }
    \label{fig:chunk-size}
\end{wrapfigure}

An important variable in \name{} is the chunk size, i.e., the length of text generated by the teacher before the student intervenes to select candidates. To examine its impact, we vary the chunk size from 512 tokens, doubling each time up to the maximum generation length of 16K tokens. We use Standard KD as a baseline. As shown in Figure~\ref{fig:chunk-size}, all chunk-size settings consistently outperform the standard KD, demonstrating the stability of \name{}. The best performance is achieved at 4K tokens. Accordingly, we adopt 4K as the default configuration in our main experiments.

\subsection{Effect of Teacher Model Size}

We further investigate whether the chunk-level mechanism in \name{} provides consistent improvements as the teacher model size increases. 
For this study, we adopt the 7B/14B/32B models from DeepSeek-R1-Distill-Qwen~\citep{deepseekai2025deepseekr1incentivizingreasoningcapability} (hereby referred to as R1-7B/14B/32B) as teacher models, both distilled from DeepSeek-R1, while keeping the student fixed as Qwen2.5-Math-1.5B.
We configure \name{} with a chunk size of 4K and a maximum generation length of 16K tokens. We use Standard KD as a baseline.

\begin{figure}[t]
    \centering
    \begin{minipage}[b]{0.49\linewidth}
        \centering
        \includegraphics[width=\linewidth]{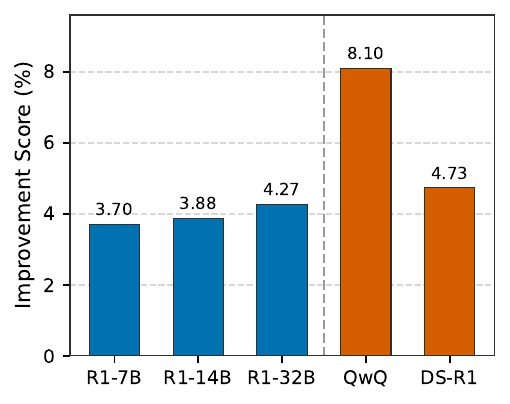}
        \caption{Performance improvements of \name{} over Standard KD across different teacher models, where the y-axis represents the average gain on Table~\ref{tab:main_exp} tasks.}
        \label{fig:teacher-comparison}
    \end{minipage}
    \hfill
    \begin{minipage}[b]{0.49\linewidth}
        \centering
        \includegraphics[width=\linewidth]{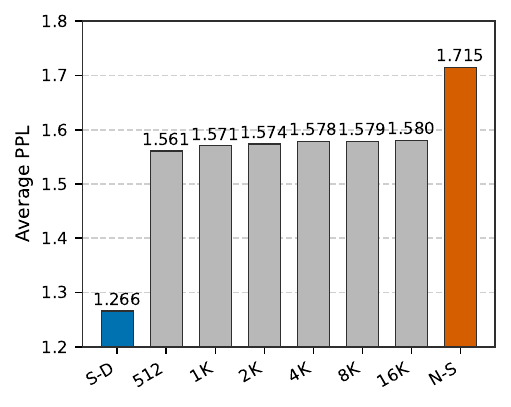}
        \caption{
        % Comparison of the average PPL of training datasets using different data generation methods. S-D: Self-Distillation. N-S: No data Selection. The middle values correspond to SSD under different chunk sizes.
        Average PPL of training data under different generation methods. S-D: Self-Distillation; N-S: No Selection; middle values: \name{} with different chunk sizes.}
        \label{fig:ppl-setting}
    \end{minipage}
\end{figure}

The results are summarized in Table~\ref{tab:R1-7B_exp}. In addition, we collect the performance of QwQ-32B and DeepSeek-R1 from Table~\ref{tab:main_exp} and ~\ref{tab:R1_exp}, compute the performance gains achieved by \name{} over Standard KD, and summarize the results in Figure~\ref{fig:teacher-comparison}, which yield the following insights:
    (1) \textbf{Scaling behavior within a unified model family}. First, within a unified model architecture, the performance gains brought by \name{} increase as the teacher model size grows: R1-32B consistently outperforms R1-14B and R1-7B. Moreover, across all teacher models, \name{} yields larger improvements than Standard KD, reaffirming the robustness of our method.
    (2) \textbf{Parameter count is not a reliable capacity indicator}. In contrast, when using QwQ-32B as the teacher, \name{} achieves lower performance compared to R1-32B despite having a similar parameter scale. Furthermore, the \name{} performance with R1-32B also surpasses that with DeepSeek-R1. These observations suggest that model parameter count alone is not an appropriate reference for comparing reasoning distillation performance across different model families.
    % We believe this phenomenon warrants further investigation in future work.

\subsection{Validity of Low-PPL Data Selection}
First, we construct training datasets based on low, high, and random PPL selections and train the student model on each. As shown in Table~\ref{tab:low-high-random-ppl_exp}, the low-PPL subset consistently yields significantly better performance than both the high-PPL and random subsets, demonstrating that selecting low-PPL data facilitates more effective learning for the student. 

% Moreover, in \name{}, the student model selects low-PPL continuations at each chunk during the teacher's generation. A natural question arises: \emph{is low-PPL data always better?} To investigate this, we compute the average PPL of the training sets used in different experiments, as shown in Figure~\ref{fig:ppl-setting}.
Moreover, in \name{}, the student model selects low-PPL continuations at each chunk during the teacher's generation. This design encourages the model to focus on reasoning steps that are more accessible to the student. A natural question then arises: \emph{is low-PPL data always better?} To examine this, we compute the average PPL of the training sets under different selection strategies, as shown in Figure~\ref{fig:ppl-setting}.

The results reveal several insights:
% \begin{itemize}
    % (1) Self-Distillation exhibits the lowest average PPL, since its training data are generated by the student itself.
    % (2) Standard KD has the highest average PPL, reflecting the distribution gap between teacher and student that prevents the student from effectively learning teacher-generated reasoning trajectories.
    % (3) \name{} lies between the two, with its average PPL gradually increasing as chunk size grows.
    (1) Self-Distillation exhibits the lowest average PPL, as the training data are generated by the student itself. While this makes the data easy to fit, it also introduces limited new learning signals. As a result, the supervision lacks diversity and may restrict the model’s ability to improve beyond its current level.
    (2) Standard KD has the highest average PPL, indicating that many teacher-generated trajectories are not well suited for the student, making them difficult to learn from effectively.
    (3) \name{} lies between the two, with its average PPL gradually increasing as the chunk size grows, reflecting a balance between learnability and informative supervision.
% \end{itemize}

% \begin{table*}
%   \centering
%   \small
%   \begin{tabular}{l|cccccc|c}
%     \toprule
%     Setting & AIME25 & AIME24 & AMC23 & OlympiadB & MATH500 & GSM8K & Avg $\uparrow$ \\
%     \midrule
%     High PPL & 7.08 & 4.37 & 34.06 & 25.82 & 59.60 & 68.39 & 33.22 \\
%     Random PPL & 9.58 & 7.92 & 35.78 & 28.64 & 65.20 & 74.75 & 36.98 \\
%     Low PPL & \textbf{10.00} & \textbf{10.21} & \textbf{41.88} & \textbf{30.71} & \textbf{67.60} & \textbf{81.88} & \textbf{40.38} \\
%     \bottomrule
%   \end{tabular}
%   \caption{Comparison of three PPL-based data selection strategies, demonstrating the superiority of low-PPL data. The experimental setting follows Table~\ref{tab:main_exp}.}
%   \label{tab:low-high-random-ppl_exp}
% \end{table*}
\begin{table*}
  \centering
  \small
  \begin{tabular}{l|ccccc|c}
    \toprule
    Setting & AIME25 & AIME24 & AMC23 & OlympiadBench & GSM8K & Avg $\uparrow$ \\
    \midrule
    High PPL & 7.08 & 4.37 & 34.06 & 25.82 & 68.39 & 27.94 \\
    Random PPL & 9.58 & 7.92 & 35.78 & 28.64 & 74.75 & 31.33 \\
    Low PPL & \textbf{10.00} & \textbf{10.21} & \textbf{41.88} & \textbf{30.71} & \textbf{81.88} & \textbf{34.94} \\
    \bottomrule
  \end{tabular}
  \caption{Comparison of three PPL-based data selection strategies, demonstrating the superiority of low-PPL data. The experimental setting follows Table~\ref{tab:main_exp}.}
  \label{tab:low-high-random-ppl_exp}
\end{table*}

These observations suggest that blindly minimizing PPL does not necessarily yield the best distillation performance. Instead, the key lies in enabling the student to actively select suitable data from the stronger teacher's sampling space, balancing learnability with reasoning richness.
Since the teacher explores a broader reasoning space, the student needs to focus on the trajectories that it can effectively learn from, rather than imitating all of them.

% \begin{figure}[htbp]
%     \centering
%     \includegraphics[width=0.5\linewidth]{pics/ppl-setting.pdf}
%     \caption{Comparison of the average PPL of training datasets using different data generation methods. S-D: Self-Distillation. N-S: No data Selection. The middle values correspond to SSD under different chunk sizes.}
%     \label{fig:ppl-setting}
% \end{figure}

\section{Conclusion and Future Work}
This paper proposes \name{}, a framework for reasoning distillation that actively involves the student model during the teacher’s sampling process. \name{} enables the student to evaluate intermediate candidates generated by the teacher using PPL, retaining reasoning trajectories that are aligned with the student’s learning capacity. 
The selected data are further refined via rejection sampling before SFT, resulting in a simple yet effective distillation pipeline.
Extensive experiments across a range of mathematical reasoning benchmarks demonstrate the effectiveness and robustness of \name{}. Compared with multiple baseline methods, \name{} consistently achieves stable performance improvements. 

Looking forward, we plan to explore two promising directions. First, we will extend \name{} to larger-scale and multimodal teacher models and apply it to additional reasoning-intensive domains, such as scientific problem solving and code generation, to further broaden its applicability. Second, we will integrate \name{} with reinforcement learning–based data selection strategies to more effectively enhance the student model’s acquisition of complex reasoning skills in a cost-efficient manner.

\section*{Ethics Statement}
All datasets used in this work (e.g., GSM8K, OlympiadBench) are publicly available and do not contain private or sensitive information. Our proposed method focuses on mathematical reasoning tasks and is not intended for direct deployment in high-stakes applications such as healthcare or law, where erroneous outputs may cause harm. While the approach involves large teacher models for data generation, \name{} reduces redundant sampling and improves efficiency, thereby mitigating excessive computational cost and environmental impact. Additionally, we used AI-assisted tools for grammar refinement in a responsible manner.
% Authors can add an optional ethics statement to the paper. 
% For papers that touch on ethical issues, this section will be evaluated as part of the review process. The ethics statement should come at the end of the paper. It does not count toward the page limit, but should not be more than 1 page. 

\bibliography{colm2026_conference}

@misc{deepseekai2025deepseekr1incentivizingreasoningcapability,
      title={DeepSeek-R1: Incentivizing Reasoning Capability in LLMs via Reinforcement Learning}, 
      author={DeepSeek-AI},
      year={2025},
      eprint={2501.12948},
      archivePrefix={arXiv},
      primaryClass={cs.CL},
      url={https://arxiv.org/abs/2501.12948}, 
}

@misc{openai2024openaio1card,
      title={OpenAI o1 System Card}, 
      author={OpenAI},
      year={2024},
      eprint={2412.16720},
      archivePrefix={arXiv},
      primaryClass={cs.AI},
      url={https://arxiv.org/abs/2412.16720}, 
}

@misc{qwq32b,
    title = {QwQ-32B: Embracing the Power of Reinforcement Learning},
    url = {https://qwenlm.github.io/blog/qwq-32b/},
    author = {Qwen Team},
    month = {March},
    year = {2025}
}

@misc{li202512surveyreasoning,
      title={From System 1 to System 2: A Survey of Reasoning Large Language Models}, 
      author={Zhong-Zhi Li and Duzhen Zhang and Ming-Liang Zhang and Jiaxin Zhang and Zengyan Liu and Yuxuan Yao and Haotian Xu and Junhao Zheng and Pei-Jie Wang and Xiuyi Chen and Yingying Zhang and Fei Yin and Jiahua Dong and Zhiwei Li and Bao-Long Bi and Ling-Rui Mei and Junfeng Fang and Xiao Liang and Zhijiang Guo and Le Song and Cheng-Lin Liu},
      year={2025},
      eprint={2502.17419},
      archivePrefix={arXiv},
      primaryClass={cs.AI},
      url={https://arxiv.org/abs/2502.17419}, 
}

@article{wei2022chain,
  title={Chain-of-thought prompting elicits reasoning in large language models},
  author={Wei, Jason and Wang, Xuezhi and Schuurmans, Dale and Bosma, Maarten and Xia, Fei and Chi, Ed and Le, Quoc V and Zhou, Denny and others},
  journal={Advances in neural information processing systems},
  volume={35},
  pages={24824--24837},
  year={2022}
}

@inproceedings{lightman2023let,
  title={Let's verify step by step},
  author={Lightman, Hunter and Kosaraju, Vineet and Burda, Yuri and Edwards, Harrison and Baker, Bowen and Lee, Teddy and Leike, Jan and Schulman, John and Sutskever, Ilya and Cobbe, Karl},
  booktitle={The Twelfth International Conference on Learning Representations},
  year={2023}
}

@inproceedings{besta2024graph,
  title={Graph of thoughts: Solving elaborate problems with large language models},
  author={Besta, Maciej and Blach, Nils and Kubicek, Ales and Gerstenberger, Robert and Podstawski, Michal and Gianinazzi, Lukas and Gajda, Joanna and Lehmann, Tomasz and Niewiadomski, Hubert and Nyczyk, Piotr and others},
  booktitle={Proceedings of the AAAI conference on artificial intelligence},
  volume={38},
  pages={17682--17690},
  year={2024}
}

@article{muennighoff2025s1,
  title={s1: Simple test-time scaling},
  author={Muennighoff, Niklas and Yang, Zitong and Shi, Weijia and Li, Xiang Lisa and Fei-Fei, Li and Hajishirzi, Hannaneh and Zettlemoyer, Luke and Liang, Percy and Cand{\`e}s, Emmanuel and Hashimoto, Tatsunori},
  journal={arXiv preprint arXiv:2501.19393},
  year={2025}
}

@article{jiang2024survey,
  title={A survey on large language models for code generation},
  author={Jiang, Juyong and Wang, Fan and Shen, Jiasi and Kim, Sungju and Kim, Sunghun},
  journal={arXiv preprint arXiv:2406.00515},
  year={2024}
}

@article{hinton2015distilling,
  title={Distilling the knowledge in a neural network},
  author={Hinton, Geoffrey and Vinyals, Oriol and Dean, Jeff},
  journal={arXiv preprint arXiv:1503.02531},
  year={2015}
}

@article{yin2025towards,
  title={Towards widening the distillation bottleneck for reasoning models},
  author={Yin, Huifeng and Zhao, Yu and Wu, Minghao and Ni, Xuanfan and Zeng, Bo and Wang, Hao and Shi, Tianqi and Shao, Liangying and Lyu, Chenyang and Wang, Longyue and others},
  journal={arXiv e-prints},
  pages={arXiv--2503},
  year={2025}
}

@inproceedings{li-etal-2025-small-models,
    title = "Small Models Struggle to Learn from Strong Reasoners",
    author = "Li, Yuetai  and Yue, Xiang  and Xu, Zhangchen  and Jiang, Fengqing  and Niu, Luyao  and Lin, Bill Yuchen  and Ramasubramanian, Bhaskar  and Poovendran, Radha",
    booktitle = "Findings of the Association for Computational Linguistics: ACL 2025",
    month = jul,
    year = "2025",
    address = "Vienna, Austria",
    publisher = "Association for Computational Linguistics",
    url = "https://aclanthology.org/2025.findings-acl.1301/",
    doi = "10.18653/v1/2025.findings-acl.1301",
    pages = "25366--25394",
}

@article{ding2025micota,
  title={MiCoTA: Bridging the Learnability Gap with Intermediate CoT and Teacher Assistants},
  author={Ding, Dongyi and Wang, Tiannan and Zhu, Chenghao and Tao, Meiling and Jiang, Yuchen Eleanor and Zhou, Wangchunshu},
  journal={arXiv preprint arXiv:2507.01887},
  year={2025}
}

@article{yang2024qwen2,
  title={Qwen2. 5-math technical report: Toward mathematical expert model via self-improvement},
  author={Yang, An and Zhang, Beichen and Hui, Binyuan and Gao, Bofei and Yu, Bowen and Li, Chengpeng and Liu, Dayiheng and Tu, Jianhong and Zhou, Jingren and Lin, Junyang and others},
  journal={arXiv preprint arXiv:2409.12122},
  year={2024}
}

@article{cobbe2021training,
  title={Training verifiers to solve math word problems},
  author={Cobbe, Karl and Kosaraju, Vineet and Bavarian, Mohammad and Chen, Mark and Jun, Heewoo and Kaiser, Lukasz and Plappert, Matthias and Tworek, Jerry and Hilton, Jacob and Nakano, Reiichiro and others},
  journal={arXiv preprint arXiv:2110.14168},
  year={2021}
}

@misc{mathai2025,
  author = {Math-AI},
  title = {Aime 2025},
  url = {https://huggingface.co/datasets/math-ai/aime25},
  year = {2025}
}

@misc{mathai2024,
  author = {Math-AI},
  title = {Aime 2024},
  url = {https://huggingface.co/datasets/math-ai/aime24},
  year = {2024}
}

@inproceedings{he-etal-2024-olympiadbench,
    title = "{O}lympiad{B}ench: A Challenging Benchmark for Promoting {AGI} with Olympiad-Level Bilingual Multimodal Scientific Problems",
    author = "He, Chaoqun  and
      Luo, Renjie  and
      Bai, Yuzhuo  and
      Hu, Shengding  and
      Thai, Zhen Leng  and
      Shen, Junhao  and
      Hu, Jinyi  and
      Han, Xu  and
      Huang, Yujie  and
      Zhang, Yuxiang  and
      Liu, Jie  and
      Qi, Lei  and
      Liu, Zhiyuan  and
      Sun, Maosong",
    booktitle = "Proceedings of the 62nd Annual Meeting of the Association for Computational Linguistics (Volume 1: Long Papers)",
    month = aug,
    year = "2024",
    address = "Bangkok, Thailand",
    publisher = "Association for Computational Linguistics",
    url = "https://aclanthology.org/2024.acl-long.211/",
    doi = "10.18653/v1/2024.acl-long.211",
    pages = "3828--3850",
}

@article{moshkov2025aimo2,
  title   = {AIMO-2 Winning Solution: Building State-of-the-Art Mathematical Reasoning Models with OpenMathReasoning dataset},
  author  = {Ivan Moshkov and Darragh Hanley and Ivan Sorokin and Shubham Toshniwal and Christof Henkel and Benedikt Schifferer and Wei Du and Igor Gitman},
  year    = {2025},
  journal = {arXiv preprint arXiv:2504.16891}
}

@misc{qwen2.5,
    title = {Qwen2.5: A Party of Foundation Models},
    url = {https://qwenlm.github.io/blog/qwen2.5/},
    author = {Qwen Team},
    month = {September},
    year = {2024}
}

@article{ye2025limo,
  title={Limo: Less is more for reasoning},
  author={Ye, Yixin and Huang, Zhen and Xiao, Yang and Chern, Ethan and Xia, Shijie and Liu, Pengfei},
  journal={arXiv preprint arXiv:2502.03387},
  year={2025}
}

@article{gu2023minillm,
  title={Minillm: Knowledge distillation of large language models},
  author={Gu, Yuxian and Dong, Li and Wei, Furu and Huang, Minlie},
  journal={arXiv preprint arXiv:2306.08543},
  year={2023}
}

@inproceedings{beyer2022knowledge,
  title={Knowledge distillation: A good teacher is patient and consistent},
  author={Beyer, Lucas and Zhai, Xiaohua and Royer, Am{\'e}lie and Markeeva, Larisa and Anil, Rohan and Kolesnikov, Alexander},
  booktitle={Proceedings of the IEEE/CVF conference on computer vision and pattern recognition},
  pages={10925--10934},
  year={2022}
}

@article{sanh2019distilbert,
  title={DistilBERT, a distilled version of BERT: smaller, faster, cheaper and lighter},
  author={Sanh, Victor and Debut, Lysandre and Chaumond, Julien and Wolf, Thomas},
  journal={arXiv preprint arXiv:1910.01108},
  year={2019}
}

@inproceedings{kim2016sequence,
  title={Sequence-level knowledge distillation},
  author={Kim, Yoon and Rush, Alexander M},
  booktitle={Proceedings of the 2016 conference on empirical methods in natural language processing},
  pages={1317--1327},
  year={2016}
}

@misc{taori2023stanford,
  title={Stanford alpaca: An instruction-following llama model},
  author={Taori, Rohan and Gulrajani, Ishaan and Zhang, Tianyi and Dubois, Yann and Li, Xuechen and Guestrin, Carlos and Liang, Percy and Hashimoto, Tatsunori B},
  year={2023},
  publisher={Stanford, CA, USA}
}

@inproceedings{wang-etal-2023-self-instruct,
    title = "Self-Instruct: Aligning Language Models with Self-Generated Instructions",
    author = "Wang, Yizhong  and Kordi, Yeganeh  and
      Mishra, Swaroop  and Liu, Alisa  and Smith, Noah A.  and Khashabi, Daniel  and Hajishirzi, Hannaneh",
    booktitle = "Proceedings of the 61st Annual Meeting of the Association for Computational Linguistics (Volume 1: Long Papers)",
    month = jul,
    year = "2023",
    address = "Toronto, Canada",
    publisher = "Association for Computational Linguistics",
    url = "https://aclanthology.org/2023.acl-long.754/",
    doi = "10.18653/v1/2023.acl-long.754",
    pages = "13484--13508",
}

@article{yang2024survey,
  title={Survey on knowledge distillation for large language models: methods, evaluation, and application},
  author={Yang, Chuanpeng and Zhu, Yao and Lu, Wang and Wang, Yidong and Chen, Qian and Gao, Chenlong and Yan, Bingjie and Chen, Yiqiang},
  journal={ACM Transactions on Intelligent Systems and Technology},
  year={2024},
  publisher={ACM New York, NY}
}

@article{faceopen,
  title={Open r1: A fully open reproduction of deepseek-r1, January 2025},
  author={Face, Hugging},
  journal={URL https://github. com/huggingface/open-r1},
  pages={9},
  year={2025},
}

@inproceedings{ho2023large,
  title={Large language models are reasoning teachers},
  author={Ho, Namgyu and Schmid, Laura and Yun, Se-Young},
  booktitle={Proceedings of the 61st annual meeting of the association for computational linguistics (volume 1: long papers)},
  pages={14852--14882},
  year={2023}
}

@misc{yan2025efficientcotdistillationselfguided,
      title={Towards Efficient CoT Distillation: Self-Guided Rationale Selector for Better Performance with Fewer Rationales}, 
      author={Jianzhi Yan and Le Liu and Youcheng Pan and Shiwei Chen and Yang Xiang and Buzhou Tang},
      year={2025},
      eprint={2509.23574},
      archivePrefix={arXiv},
      primaryClass={cs.CL},
      url={https://arxiv.org/abs/2509.23574}, 
}

@article{chen2023mcc,
  title={Mcc-kd: Multi-cot consistent knowledge distillation},
  author={Chen, Hongzhan and Wu, Siyue and Quan, Xiaojun and Wang, Rui and Yan, Ming and Zhang, Ji},
  journal={arXiv preprint arXiv:2310.14747},
  year={2023}
}

@article{zhang2024elad,
  title={Elad: Explanation-guided large language models active distillation},
  author={Zhang, Yifei and Pan, Bo and Ling, Chen and Hu, Yuntong and Zhao, Liang},
  journal={arXiv preprint arXiv:2402.13098},
  year={2024}
}

@article{zhou2024teaching,
  title={Teaching-assistant-in-the-loop: Improving knowledge distillation from imperfect teacher models in low-budget scenarios},
  author={Zhou, Yuhang and Ai, Wei},
  journal={arXiv preprint arXiv:2406.05322},
  year={2024}
}

@article{austin2021program,
  title={Program synthesis with large language models},
  author={Austin, Jacob and Odena, Augustus and Nye, Maxwell and Bosma, Maarten and Michalewski, Henryk and Dohan, David and Jiang, Ellen and Cai, Carrie and Terry, Michael and Le, Quoc and others},
  journal={arXiv preprint arXiv:2108.07732},
  year={2021}
}

@article{clark2018think,
  title={Think you have solved question answering? try arc, the ai2 reasoning challenge},
  author={Clark, Peter and Cowhey, Isaac and Etzioni, Oren and Khot, Tushar and Sabharwal, Ashish and Schoenick, Carissa and Tafjord, Oyvind},
  journal={arXiv preprint arXiv:1803.05457},
  year={2018}
}

@article{geva-etal-2021-aristotle,
    title = "Did Aristotle Use a Laptop? A Question Answering Benchmark with Implicit Reasoning Strategies",
    author = "Geva, Mor  and Khashabi, Daniel  and Segal, Elad  and Khot, Tushar  and Roth, Dan  and Berant, Jonathan",
    journal = "Transactions of the Association for Computational Linguistics",
    volume = "9",
    year = "2021",
    address = "Cambridge, MA",
    publisher = "MIT Press",
    url = "https://aclanthology.org/2021.tacl-1.21/",
    pages = "346--361",
}

@inproceedings{he-etal-2024-ultraeval,
    title = "{U}ltra{E}val: A Lightweight Platform for Flexible and Comprehensive Evaluation for {LLM}s",
    author = "He, Chaoqun  and
      Luo, Renjie  and
      Hu, Shengding  and
      Zhao, Ranchi  and
      Zhou, Jie  and
      Wu, Hanghao  and
      Zhang, Jiajie  and
      Han, Xu  and
      Liu, Zhiyuan  and
      Sun, Maosong",
    editor = "Cao, Yixin  and
      Feng, Yang  and
      Xiong, Deyi",
    booktitle = "Proceedings of the 62nd Annual Meeting of the Association for Computational Linguistics (Volume 3: System Demonstrations)",
    month = aug,
    year = "2024",
    address = "Bangkok, Thailand",
    publisher = "Association for Computational Linguistics",
    url = "https://aclanthology.org/2024.acl-demos.23/",
    doi = "10.18653/v1/2024.acl-demos.23",
    pages = "247--257"
}

@inproceedings{talmor-etal-2019-commonsenseqa,
    title = "{C}ommonsense{QA}: A Question Answering Challenge Targeting Commonsense Knowledge",
    author = "Talmor, Alon  and
      Herzig, Jonathan  and
      Lourie, Nicholas  and
      Berant, Jonathan",
    editor = "Burstein, Jill  and
      Doran, Christy  and
      Solorio, Thamar",
    booktitle = "Proceedings of the 2019 Conference of the North {A}merican Chapter of the Association for Computational Linguistics: Human Language Technologies, Volume 1 (Long and Short Papers)",
    month = jun,
    year = "2019",
    address = "Minneapolis, Minnesota",
    publisher = "Association for Computational Linguistics",
    url = "https://aclanthology.org/N19-1421/",
    doi = "10.18653/v1/N19-1421",
    pages = "4149--4158"
}

@article{lu2025onpolicydistillation,
  author = {Kevin Lu and Thinking Machines Lab},
  title = {On-Policy Distillation},
  journal = {Thinking Machines Lab: Connectionism},
  year = {2025},
  note = {https://thinkingmachines.ai/blog/on-policy-distillation},
  doi = {10.64434/tml.20251026},
}

@article{yang2025qwen3,
  title={Qwen3 technical report},
  author={Yang, An and Li, Anfeng and Yang, Baosong and Zhang, Beichen and Hui, Binyuan and Zheng, Bo and Yu, Bowen and Gao, Chang and Huang, Chengen and Lv, Chenxu and others},
  journal={arXiv preprint arXiv:2505.09388},
  year={2025}
}

@misc{mimo2025flash,
  title={MiMo-V2-Flash Technical Report},
  author={LLM-Core Xiaomi},
  year={2025},
  url={https://github.com/XiaomiMiMo/MiMo-V2-Flash/paper.pdf}
}

@inproceedings{agarwal2024policy,
  title={On-policy distillation of language models: Learning from self-generated mistakes},
  author={Agarwal, Rishabh and Vieillard, Nino and Zhou, Yongchao and Stanczyk, Piotr and Garea, Sabela Ramos and Geist, Matthieu and Bachem, Olivier},
  booktitle={The Twelfth International Conference on Learning Representations},
  year={2024}
}
\bibliographystyle{colm2026_conference}

\appendix
% \section{Appendix}

\section{Implementation Details}

During the data selection, to ensure data diversity and enlarge the exploration space, we retain the two candidates with the lowest PPL at each chunk after student evaluation. Then, at the end of sampling, only the single candidate with the lowest overall PPL is preserved.

For the cold-start stage, we use 3K training samples. In the main distillation stage, both Standard KD and \name{} are trained on 8.5K samples for open-source models, while 10K samples are used for closed-source models. 
We reproduce MCC-KD and MoRSD by strictly following the methodological descriptions in their respective papers. For MoRSD on mathematical reasoning tasks, we adopt $\delta$ = 1 in the accuracy selection stage and set K = 5 for diversity selection.
The hyperparameters used in SFT as illustrated in Table~\ref{tab:hyper-setting}. 
% The method is summarized in Algorithm~\ref{alg:ssd}.

\begin{algorithm}[t]
\caption{Student-guided sampling with PPL-based selection during generation}
\label{alg:ssd}
\begin{algorithmic}[1]
\REQUIRE Teacher model $T$, Student model $S$, input $x$, 
number of candidates $K$, max steps $L$
\ENSURE Simplified reasoning path $R$

\STATE $R \leftarrow \emptyset$
\STATE $context \leftarrow x$

\FOR{$t = 1$ to $L$}

    \STATE Sample candidate chunks:
    \STATE $c_1, c_2, \dots, c_K \sim T(\cdot \mid context)$

    \STATE Evaluate candidates using student model:
    \FOR{$k = 1$ to $K$}
        \STATE $score_k \leftarrow \text{Evaluate}(S, context, c_k)$
    \ENDFOR

    \STATE Select best chunk:
    \STATE $c^* \leftarrow \arg\min_k score_k$

    \STATE Update reasoning path:
    \STATE $R \leftarrow R \cup \{c^*\}$
    \STATE $context \leftarrow context \oplus c^*$

    \IF{StopCondition$(context)$}
        \STATE \textbf{break}
    \ENDIF

\ENDFOR

\RETURN $R$
\end{algorithmic}
\end{algorithm}

\begin{table}
  \centering
  \renewcommand{\arraystretch}{1.1}
  \begin{tabular}{lc}
    \toprule
    \textbf{Hyperparameter} & \textbf{Value}\\
    \midrule
    Learning Rate & $2 \times 10^{-5}$ \\ 
    Number of Epochs & 5 \\
    Number of Devices & 4 \\
    Batch Size & 32 \\
    % Gradient Accumulation Steps & 8 \\
    Optimizer & AdamW \\
    Learning Rate Scheduler & cosine \\
    Weight Decay & 0.01 \\
    Warmup Steps & 50 \\
    Max Sequence Length & 16,384 \\
    \bottomrule
  \end{tabular}
  \caption{The hyperparameters used for fine-tuning.}
  \label{tab:hyper-setting}
\end{table}

% \section{Example Appendix}
% \label{sec:appendix}

\section{Evaluation Benchmarks}
Here, we briefly introduce the five benchmarks used for evaluating mathematical reasoning in the main experiments.
\begin{itemize}
\item \textbf{AIME24/25} consists of problems from the American Invitational Mathematics Examination (AIME), which focus on high-difficulty competition-level mathematical reasoning. Each benchmark contains 30 problems.

\item \textbf{AMC23} is derived from the American Mathematics Competition (AMC) and targets intermediate-level mathematical reasoning. The evaluation subset contains 40 problems.

\item \textbf{OlympiadBench}~\citep{he-etal-2024-olympiadbench} is a challenging benchmark designed to assess Olympiad-level mathematical reasoning. We use the English Mathematical Olympiad subset containing 674 problems.

% \item \textbf{MATH500}~\citep{hendrycks2021measuring} is a curated subset of the MATH dataset that focuses on advanced mathematical reasoning. The benchmark consists of 500 problems.

\item \textbf{GSM8K}~\citep{cobbe2021training} is a widely used benchmark for grade-school mathematical word problem solving. The problems require multi-step arithmetic and logical reasoning. The test set contains 1,319 problems.

% \textbf{MBPP} evaluates code generation and program reasoning capabilities. This benchmark contains approximately 1,000 crowd-sourced Python programming problems.

% \textbf{ARC-C} (AI2 Reasoning Challenge – Challenge Set) focuses on scientific and commonsense reasoning in a multiple-choice format. The questions are designed to be challenging, often requiring abstract reasoning and the integration of multiple pieces of knowledge. The benchmark includes 1,172 test problems.

% \textbf{StrategyQA} is a benchmark for multi-hop factual and strategic reasoning, formulated as yes/no questions. Answering these questions typically requires implicit reasoning over multiple facts rather than relying on a single piece of information. The test set contains 2,290 problems.

% \textbf{CommonsenseQA} is a multiple-choice benchmark constructed from ConceptNet to evaluate commonsense reasoning. It emphasizes everyday knowledge and semantic associations that are not explicitly stated in text. The test set contains 1,221 problems.
\end{itemize}

\section{Effect of Progressive Sampling Reduction}
\label{app:sampling_decrease}

To reduce computational cost during sampling, we progressively decrease the number of candidates. We design controlled experiments to validate the effectiveness of this strategy. Specifically, we conduct two sets of experiments under chunk sizes of 1K and 4K, respectively. In each setting, we compare (i) a fixed sampling strategy that maintains 16 candidates throughout the generation process and (ii) a progressively decreasing sampling strategy.

Following the same data construction and training pipeline, we evaluate the resulting models on Benchmarks and measure the time required to sample 500 examples. As shown in Table~\ref{tab:reduction_comparison}, for both 1K and 4K chunk sizes, the decreasing sampling strategy substantially reduces sampling time while preserving stable performance. Notably, it does not lead to performance degradation and even yields slight improvements. These results indicate that progressively reducing the sampling budget does not compromise the quality of previously selected low-ppl reasoning trajectories, thereby maintaining overall data quality.

% \begin{table}
%   \centering
%   \small
%   \renewcommand{\arraystretch}{1.2}
%   \begin{tabular}{l|cc}
%     \toprule
%     \textbf{Setting} & \textbf{Time}$\downarrow$ & \textbf{Accuracy}$\uparrow$\\
%     \midrule
%     \textit{1K} & & \\
%     \hspace{1em} 16 candidates & 35.46 & 36.68 \\ 
%     \hspace{1em} Decreasing sampling & \textbf{28.20} & \textbf{37.58} \\
%     \hline
%     \textit{4K} & & \\
%     \hspace{1em} 16 candidates & 34.22 & 38.95 \\
%     \hspace{1em} Decreasing sampling & \textbf{31.59} & \textbf{40.38} \\
%     \bottomrule
%   \end{tabular}
%   \caption{Comparison results. Accuracy is computed as the average performance over the six benchmarks in Table~\ref{tab:main_exp}. The best scores are bolded.}
%   \label{tab:reduction_comparison}
% \end{table}
\begin{table}
  \centering
  \small
  \renewcommand{\arraystretch}{1.2}
  \begin{tabular}{l|cc}
    \toprule
    \textbf{Setting} & \textbf{Time}$\downarrow$ & \textbf{Accuracy}$\uparrow$\\
    \midrule
    \textit{1K} & & \\
    \hspace{1em} 16 candidates & 35.46 & 31.22 \\ 
    \hspace{1em} Decreasing sampling & \textbf{28.20} & \textbf{31.73} \\
    \hline
    \textit{4K} & & \\
    \hspace{1em} 16 candidates & 34.22 & 33.66 \\
    \hspace{1em} Decreasing sampling & \textbf{31.59} & \textbf{34.94} \\
    \bottomrule
  \end{tabular}
  \caption{Comparison results. Accuracy is computed as the average performance over the five benchmarks in Table~\ref{tab:main_exp}. The best scores are bolded.}
  \label{tab:reduction_comparison}
\end{table}

\section{Application to Closed-Source Models}
\label{sec:application-to-closed-source}

Many of the strongest reasoning models are closed-source and only accessible through APIs. In such settings, users can only provide prompts but cannot intervene in the sampling process. For example, DeepSeek does not allow modifying parameters such as temperature\footnote{\url{https://api-docs.deepseek.com/guides/reasoning_model}}. These constraints directly affect the applicability of \name{}. Moreover, most APIs prepend an immutable system prompt, which may cause misalignment during chunk concatenation.

To mitigate these issues, and given that our later experiments (Section~\ref{sec:chunk-size}) showed only marginal differences between 4K and 16K chunk sizes, we fix the chunk size to 16K in this section for evaluating \name{} under closed-source conditions.

For this experiment, we use the R1 portion of the OpenMathReasoning dataset, where multiple solutions are generated by DeepSeek-R1. We filter for problems with 16 correct and available solutions, consistent with our main experimental setup. This simulates the \name{} process under a chunk size of 16K. Note that in practice, the actual number of sampled solutions may exceed 16. Nevertheless, we still apply the student model to select the candidate with the lowest PPL for training.

We compare two baselines in this setting:
% \begin{itemize}
    (1) \textbf{Standard KD.} Distillation directly from teacher outputs. Since the data are already pre-generated, we simulate this condition by fixing the random seed to 35 and randomly selecting one solution per problem.
    (2) \textbf{\name{}.} The student model selects the lowest-PPL solution from the teacher's candidates.
% \end{itemize}

As shown in Table~\ref{tab:R1_exp}, \name{} consistently outperforms Standard KD, both with and without cold start. In particular, \name{} achieves an average improvement of 4 points over the baseline, demonstrating that \name{} provides stable and robust gains even when applied to closed-source or API-based scenarios, highlighting its broad applicability.

% \begin{table*}
%   \centering
%   \small
%   % \renewcommand{\arraystretch}{1.2}
%   % \resizebox{\textwidth}{!}{
%   \begin{tabular}{l|cccccc|c}
%     \toprule
%     Method & AIME25 & AIME24 & AMC23 & OlympiadB & MATH500 & GSM8K & Avg $\uparrow$ \\
%     % \textbf{Method}\rule{0pt}{3ex} & \textbf{AIME25} & \textbf{AIME24} & \textbf{AMC23} & \makecell{\textbf{Olympiad} \\ \textbf{Bench}} & \textbf{MATH500} & \textbf{GSM8K} & \textbf{Average} $\uparrow$\\
%     \midrule
%     Standard KD* & 7.92 & 6.46 & 38.12 & 26.85 & 60.60 & 78.24 & 36.37  \\
%     \name{}* & \textbf{10.00} & 6.88 & 44.37 & 31.01 & \textbf{71.00} & 82.26 & 40.92 \\
%     \midrule
%     Cold Start & 6.04 & 4.79 & 29.06 & 24.93 & 54.20 & 65.96 & 30.83\\
%     Standard KD & 9.58 & 7.50 & 37.97 & 30.56 & 64.80 & 77.41 & 37.97 \\
%     \name{} & 9.79 & \textbf{8.96} & \textbf{45.16} & \textbf{32.49} & 70.00 & \textbf{84.84} & \textbf{41.87} \\
%     \bottomrule
%   \end{tabular}
%   \caption{Performance comparison across various benchmarks with closed-source teacher models. The highest scores are bolded. Here, * indicates direct distillation on Qwen2.5-Math-1.5B without the cold-start stage.}
%   \label{tab:R1_exp}
% \end{table*}
\begin{table*}
  \centering
  \small
  % \renewcommand{\arraystretch}{1.2}
  % \resizebox{\textwidth}{!}{
  \begin{tabular}{l|ccccc|c}
    \toprule
    Method & AIME25 & AIME24 & AMC23 & OlympiadBench & GSM8K & Avg $\uparrow$ \\
    % \textbf{Method}\rule{0pt}{3ex} & \textbf{AIME25} & \textbf{AIME24} & \textbf{AMC23} & \makecell{\textbf{Olympiad} \\ \textbf{Bench}} & \textbf{MATH500} & \textbf{GSM8K} & \textbf{Average} $\uparrow$\\
    \midrule
    Standard KD* & 7.92 & 6.46 & 38.12 & 26.85 & 78.24 & 31.52  \\
    \name{}* & \textbf{10.00} & 6.88 & 44.37 & 31.01 & 82.26 & 34.90 \\
    \midrule
    Cold Start & 6.04 & 4.79 & 29.06 & 24.93 & 65.96 & 26.16\\
    Standard KD & 9.58 & 7.50 & 37.97 & 30.56 & 77.41 & 32.60 \\
    \name{} & 9.79 & \textbf{8.96} & \textbf{45.16} & \textbf{32.49} & \textbf{84.84} & \textbf{36.25} \\
    \bottomrule
  \end{tabular}
  \caption{Performance comparison across various benchmarks with closed-source teacher models. The highest scores are bolded. Here, * indicates direct distillation on Qwen2.5-Math-1.5B without the cold-start stage.}
  \label{tab:R1_exp}
\end{table*}

\begin{table*}
  \centering
  \small
  \setlength{\tabcolsep}{3pt}
  \begin{tabular}{l|cccc|c}
    \toprule
    Method & MBPP & ARC-C & StrategyQA & CommonsenseQA & Average $\uparrow$ \\
    % \textbf{Method}\rule{0pt}{3ex} & \textbf{AIME25} & \textbf{AIME24} & \textbf{AMC23} & \makecell{\textbf{Olympiad} \\ \textbf{Bench}} & \textbf{MATH500} & \textbf{GSM8K} & \textbf{Average} $\uparrow$ \\
    \midrule
    % base & & & & & & & \\
    Standard KD & 20.03 & 38.65 & 63.23 & 41.66 & 40.89 \\
    MCC-KD & 17.76 & 38.14 & 64.37 & 40.67 & 40.24 \\
    MoRSD & \textbf{22.07} & 38.91 & 64.93 & 40.43 & 41.59 \\
    \name{} & 21.36 & \textbf{39.68} & \textbf{65.68} & \textbf{42.07} & \textbf{42.20} \\
    \bottomrule
  \end{tabular}
  \caption{Evaluation results on MBPP, ARC-C, StrategyQA, and CommonsenseQA, demonstrating generalization beyond mathematical reasoning. The best scores are bolded.}
  \label{tab:generalizability_exp}
\end{table*}

\section{Generalization Evaluation}

Beyond complex mathematical reasoning tasks, we further evaluate our method on code generation (MBPP~\citep{austin2021program}) and general reasoning benchmarks (ARC-C~\citep{clark2018think}, StrategyQA~\citep{geva-etal-2021-aristotle}, and CommonsenseQA~\citep{talmor-etal-2019-commonsenseqa}) to better assess its generalization ability. We conduct these evaluations using the UltraEval framework~\citep{he-etal-2024-ultraeval} with its default configuration. As shown in Table~\ref{tab:generalizability_exp}, our method consistently outperforms the baseline approaches, demonstrating strong generalization across diverse task domains.

\begin{table*}
  \centering
  \small
  \begin{tabular}{l|ccccc|c}
    \toprule
    Method & AIME25 & AIME24 & AMC23 & OlympiadBench & GSM8K & Avg $\uparrow$ \\
    \midrule
    % base & & & & & & & \\
    Standard KD* & 4.37 & 3.75 & 33.28 & 23.29 & 69.52 & 26.84 \\ \midrule
    \textit{\name{}} & & & & & & \\
    \hspace{1em} 512 & 8.33 & 6.04 & 33.59 & 26.71 & 74.53 & 29.84\\
    \hspace{1em} 1K & 8.54 & 6.25 & 37.5 & 28.19 & 78.17 & 31.73\\
    \hspace{1em} 2K & 7.50 & 6.67 & 40.78 & 27.60 & 74.37 & 31.38\\
    \hspace{1em} 4K & 10.00 & \textbf{10.21} & \textbf{41.88} & 30.71 & \textbf{81.88} & \textbf{34.94} \\
    \hspace{1em} 8K & 9.38 & 6.46 & 38.12 & 28.19 & 76.57 & 31.74\\
    \hspace{1em} 16K & \textbf{11.25} & 7.08 & 39.22 & \textbf{31.31} & 79.53 & 33.68\\
    \bottomrule
  \end{tabular}
  \caption{Benchmarks performance of \name{} with varying chunk sizes, compared against Standard KD as the baseline. Best results are bolded. Here, * indicates direct distillation on Qwen2.5-Math-1.5B without the cold-start stage.}
  \label{tab:benchmark_exp}
\end{table*}

\section{More Experiment Results}

Table~\ref{tab:benchmark_exp} reports the performance of \name{} under different chunk sizes across all benchmarks, with Standard KD as the baseline. Table~\ref{tab:R1-7B_exp} presents the performance of our method compared with Standard KD when using R1-7B, R1-14B, and R1-32B as teacher models.

\begin{table*}
  \centering
  \small
  \begin{tabular}{l|ccccc|c}
    \toprule
    Method & AIME25 & AIME24 & AMC23 & OlympiadBench & GSM8K & Avg $\uparrow$ \\
    \midrule
    \textit{R1-7B} & & & & & & \\
    \hspace{1em} Standard KD* & 7.92 & 8.33 & 38.91 & 26.26 & 76.27 & 31.54\\
    \hspace{1em} \name{} & \textbf{10.83} & \textbf{9.79}  & \textbf{41.88} & \textbf{31.16} & \textbf{82.56} & \textbf{35.24}\\
    \midrule
    \textit{R1-14B} & & & & & & \\
    \hspace{1em} Standard KD* & 10.21 & 6.88 & 37.81 & 26.41 & 77.10 & 31.68\\
    \hspace{1em} \name{} & \textbf{12.08} & \textbf{9.17} & \textbf{45.31} & \textbf{30.71} & \textbf{80.52} & \textbf{35.56} \\
    \midrule
    \textit{R1-32B}  & & & & & & \\
    \hspace{1em} Standard KD* & 11.67 & 7.92 & 37.19 & 31.01 & 78.01 & 33.16 \\
    \hspace{1em} \name{} & \textbf{15.42} & \textbf{12.29} & \textbf{46.72} & \textbf{32.79} & \textbf{79.91} & \textbf{37.43} \\
    \bottomrule
  \end{tabular}
  \caption{Comparison of Standard KD and \name{} using R1-7B/14B/32B as teacher models. The highest scores are bolded. Here, * indicates direct distillation on Qwen2.5-Math-1.5B without the cold-start stage.}
  \label{tab:R1-7B_exp}
\end{table*}

\section{Case Study: Comparing Reasoning Patterns in a High-PPL Chunk}
\label{sec:chunk_analysis}

We further analyze the segment at index 26, where the PPL difference is particularly large. The problem considered is: ``\textit{In a class, 18 fathers and 24 mothers attended a parent meeting. Both parents of 10 male students and 8 female students, only the mother of 4 male students and 3 female students, and only the father of 1 male student and 1 female student attended the meeting. How many students are in the class?}''

\begin{figure*}[htbp]
    \centering
    \includegraphics[width=0.75\linewidth]{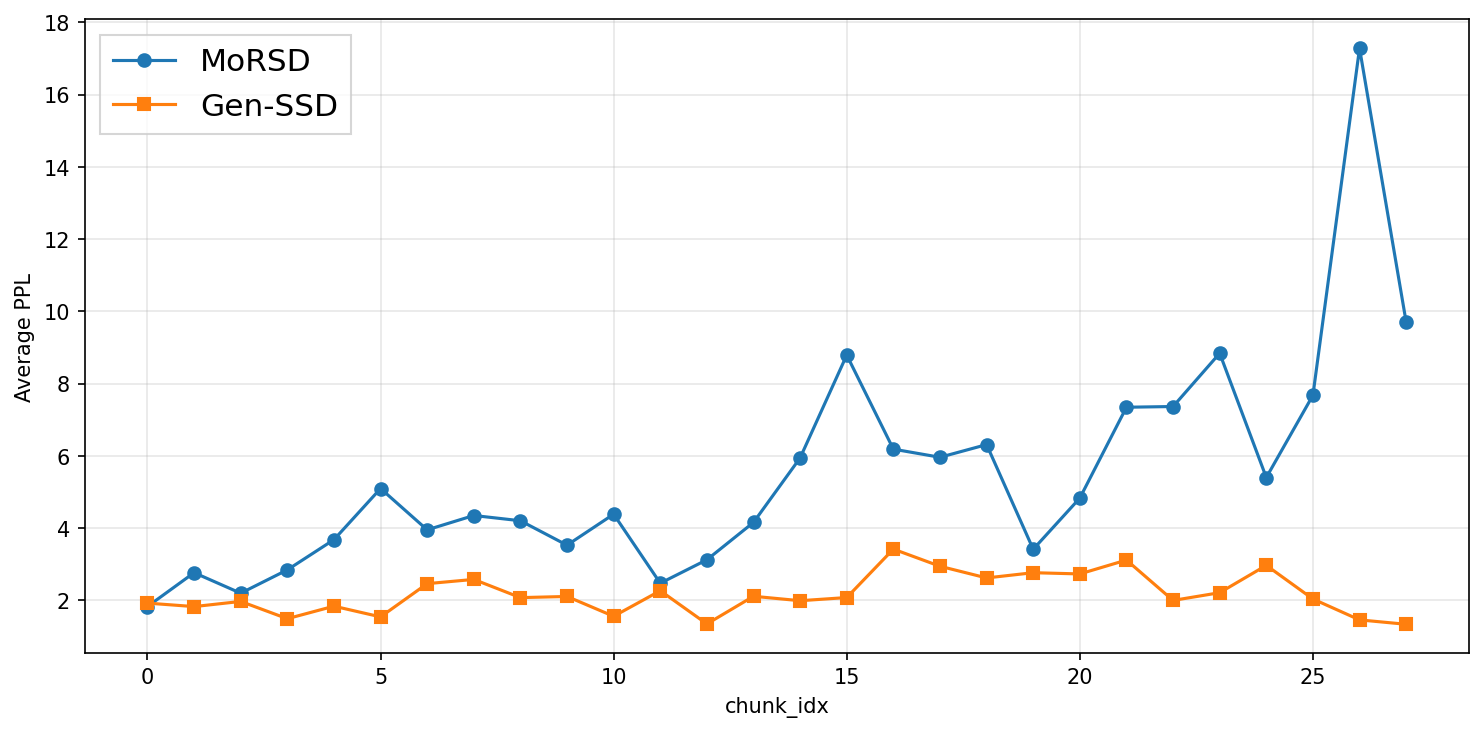}
    \caption{Comparison of PPL trends between \name{} and MoRSD.}
    \label{fig:ppl-trend}
\end{figure*}

For MoRSD, the corresponding segment exhibits unstable and repetitive reasoning, with frequent revisions and no clear convergence. In contrast, the segment selected by \name{} shows a more stable and structured reasoning process, with clearer steps and a more direct path toward the solution.

\begin{tcolorbox}[
  enhanced,
  breakable,
  colback=white,
  colframe=black,
  boxrule=0.8pt,
  arc=3pt,
  left=8pt,
  right=8pt,
  top=10pt,
  bottom=8pt,
  width=1\linewidth,
  title=Chunk from MoRSD,
  fonttitle=\bfseries,
  coltitle=black,
  boxed title style={
    colback=white,
    colframe=black,
    boxrule=0.8pt,
    arc=3pt,
    left=4pt,
    right=4pt,
    top=2pt,
    bottom=2pt
  },
  attach boxed title to top left={
    xshift=6pt,
    yshift=-3mm
  }
]
\begin{Verbatim}[breaklines=true]
the meeting."

The parents here can be either both fathers? No, both parents: father and mother of each child.

So each child has a father and a mother, so each child’s both parents would be a father and a mother.

Thus the mention of "mother of male students" means that for some male student (male child), their mother attended (maybe alone).

Therefore the student's gender is separate from their parents' gender.

Thus the counts were okay as previously.

Therefore, the problem may be designed such that students not included in these categories had no parents, so the parent numbers given must subtract the surplus from somewhere else.

Alternatively, maybe there is an overlap. Perhaps one father attended both for both a son and a daughter, so he’s being counted in two places.

Wait, but each parent is a parent of a single child?

Probably not assuming that, since parents have more children. But the problem specifies "parent meeting"— probably each child's parents are present or not as per the problem.

Alternatively, perhaps a mother with a male and female child would count twice, but unless there is data on family linkages.

Alternatively, perhaps it's expecting to assume that the given parent totals (18+24) is precisely the counts from the stated"
\end{Verbatim}
\end{tcolorbox}

\begin{tcolorbox}[
  enhanced,
  breakable,
  colback=gray!10,
  colframe=black,
  boxrule=1.2pt,
  arc=6pt,
  % ——关键：用 title，而不是 overlay——
  title={Chunk from \name{}},
  fonttitle=\bfseries\color{white},
  coltitle=white,
  boxed title style={
    colback=black,
    colframe=black,
    boxrule=0pt,
    arc=3pt,
    left=6pt,right=6pt,top=3pt,bottom=3pt
  },
  attach boxed title to top left={xshift=10pt,yshift=-2mm},
  % 给标题留出一点框内空间（避免标题压到正文）
  top=6mm,
  left=12pt,right=12pt,bottom=10pt
    ]
\begin{Verbatim}[breaklines=true]
female student

2. **Calculate the total number of students:**
- Male students: \(10 \text{ (both)} + 4 \text{ (only mother)} + 1 \text{ (only father)} = 15\)
- Female students: \(8 \text{ (both)} + 3 \text{ (only mother)} + 1 \text{ (only father)} = 12\)
- Total students: \(15 + 12 = 27\)

3. **Verify the number of fathers and mothers who attended:**
- **Fathers:**
- From both parents: \(10 \text{ (male)} + 8 \text{ (female)} = 18\)
- From only father students: 1 (male) + 1 (female) = 2
- However, to reconcile with the given 18 fathers, the 2 fathers from only father students must be among the 18 already counted.
- **Mothers:**
- From both parents: \(10 \text{ (male)} + 8 \text{ (female)} = 18\)
- From"
\end{Verbatim}
\end{tcolorbox}

% \begin{table*}
%   \centering
%   \small
%   \begin{tabular}{l|cccccc|c}
%     \toprule
%     Method & AIME25 & AIME24 & AMC23 & OlympiadBench & MATH500 & GSM8K & Average $\uparrow$ \\
%     \midrule
%     \rowcolor{gray!20}
%     Teacher(QwQ-32B) & 49.79 & 67.50 & 91.09 & 66.91 & 94.20 & 96.13 & 77.60 \\
%     \midrule
%     Test & 6.46 & 6.67 & 35.00 & 28.04 & 66.00 & 68.99 & 35.19 \\
%     \name{}(ours) & \textbf{10.00} & \textbf{8.96} & \textbf{41.88} & \textbf{30.71} & \textbf{66.40} & \textbf{79.98} & \textbf{39.66} \\
%     \bottomrule
%   \end{tabular}
%   \caption{Performance comparison across various benchmarks. We adopt QwQ-32B as the teacher model and Qwen2.5-Math-1.5B as the student model. Best results are bolded. Here, * indicates direct distillation on Qwen2.5-Math-1.5 without the cold-start stage. 
%   The chunk size for the \name{} is 4K.}
%   \label{tab:main_exp-1}
% \end{table*}

% \begin{equation}
% \begin{aligned}
% L \left(y_c^{(k)} \right) 
% = - \frac{1}{ \left\vert y_c^{(k)} \right\vert} 
%   \sum_{t=1}^{ \left \vert y_c^{(k)} \right \vert} 
%   \log P_S\big( y_{c,t}^{(k)} \mid & \\ x, y_{<c}, y_{c,<t}^{(k)} \big),
% \end{aligned}
% \end{equation}

% \begin{equation}
% \text{PPL} \left(y_c^{(k)} \right) = \exp\left( L \left(y_c^{(k)} \right) \right).
% \end{equation}

\end{document}